\documentclass[cameraready]{Interspeech}

\title{Reasoning Beyond Transcription: Audio Language Models on Child Stuttering Speech}

\author[affiliation={1}]{Chibuzor}{Okocha}
\author[affiliation={1}]{Christan}{Grant}
\author[affiliation={1}]{Zoey}{Liu}
 
\address{
    $^1$ University of Florida, USA
}

\email{c.okocha@ufl.edu, christan@ufl.edu, liu.ying@ufl.edu}
 

\usepackage{comment}
\usepackage{float}

\begin{document}
\maketitle

\begin{abstract}
Child speech differs from adult speech in acoustics, prosody, and linguistic structures. Speech disfluencies (such as repetitions) further challenge automatic understanding. While Audio–Language Models (ALMs) show strong semantic reasoning from speech audio, their ability to reason about disfluent child speech in mixed-speaker settings remains unexplored. We investigate this through two tasks: child-focused semantic summarization and speech entailment. Experiments use recordings of children who stutter in mixed-speaker interviews without explicit speaker separation. Models are instruction-guided to focus on the child, preserve clinically relevant disfluencies, and avoid adult-speech leakage. Evaluation combines LLM-based judges and reference-based metrics, anchored by transcript-oracle baselines to isolate errors. Results show that while ALMs extract high-level meaning from stuttered speech, reasoning degrades significantly with increased disfluency and speaker interference.
\end{abstract}

\section{Introduction}

Compared to adult speech, child speech is markedly different from adult speech in acoustic realization, prosody, and linguistic structures due to ongoing physiological and cognitive development \cite{gerosa2007childASR, potamianos2003children, dejong2023}. For children who stutter, these differences are further amplified by speech disfluencies such as repetitions, prolongations, and stutters \cite{lee_acoustics_1999, millager2023, arenas2017}. While these disfluencies are clinically meaningful \cite{manning2023clinical}, they pose persistent challenges for automatic speech recognition (ASR) and downstream language processing systems. Most current systems are optimized for fluent adult speech and often treat disfluency as noise to be removed \cite{potamianos2003children, lee_acoustics_1999}. This creates a transcription bottleneck where vital information pertaining to characteristics unique to child speech is lost or ``corrected" before it can be analyzed.

Recent advances in Audio--Language Models (ALMs) offer a potential solution by enabling direct processing and semantic reasoning over raw audio \cite{huang2024audiogpt, gong2024, rubenstein2023}. By bypassing explicit transcription, these models provide new opportunities for speech understanding in adverse conditions \cite{yaruss2006}. While ALMs have shown promising results in tasks like audio captioning and general summarization \cite{wang2025dualspeechlm, deshmukh}, their ability to reason about disfluent child speech is not yet established. Specifically, it remains unclear if these models can extract semantically faithful information, preserve clinically relevant disfluencies, and avoid hallucinations when faced with irregular acoustic patterns \cite{okocha2025a, okocha2026evaluating}.

This challenge is particularly acute in semi-structured interview settings where child speech is interleaved with adult prompts \cite{okocha2025a}. Classical approaches to multi-speaker audio typically rely on diarization or source separation pipelines \cite{sanni_afrispeech-dialog_2025, okocha_domain-aware_2025}. However, such signal-level processing can be brittle in the presence of high-frequency disfluency. 

In this work, we investigate whether state-of-the-art ALMs can perform child-focused semantic reasoning directly from raw audios that contain disfluent child speech. To assess both understanding and reasoning, we study two complementary tasks: (1) child-only semantic summarization, where models must generate concise summaries that capture the child's speech content while preserving disfluencies and excluding adult speech; (2) A novel child speech entailment task, addressing the absence of semantic reasoning benchmarks for disfluent child speech in audio-native models.

In summary, our work makes the following contributions. First, we formalize the problem of child-focused semantic reasoning across single- and mixed-speaker settings.  We explicitly distinguish this from classical source separation by focusing on instruction-based intent.
Second, we evaluate popular ALMs on child-only semantic summarization of disfluent  speech. Our analysis focuses on faithfulness, disfluency preservation, and robustness to adult-speech interference.
Third, our child speech entailment task provides a controlled probe of semantic reasoning under disfluency and mixed-speaker conditions with stratified difficulty levels.
We anchor our automated evaluations with an ASR and LLM cascaded baselines. This allows us to isolate sources of error and determine if performance gaps are due to audio recognition in the ASR or semantic reasoning in the LLM. Together, these tasks provide a multifaceted assessment of disfluency-aware child speech understanding in clinically and educationally relevant settings.

\section{Method and Task Definition}

Here we introduce our formalism for the problem of child-focused semantic reasoning from raw audio. Given a raw audio recording $x$ containing interleaved child and adult speech, the objective is to assess whether an ALM can extract the child's semantic intent while preserving clinically relevant disfluencies and ignoring adult-speech interference.

\subsection{Task 1: Child-Only Semantic Summarization}

Given audio $x$ and a natural language prompt $p$, the model $f(\cdot)$ must generate a concise summary $y = f(x, p)$. In mixed-speaker interviews, $p$ explicitly directs the model to focus only on the child’s speech. This task evaluates if ALMs can perform instruction-guided diarization at the semantic level without signal-level separation. 

To isolate error sources, we utilize an ASR and LLM baseline. As a strong text-based baseline, we evaluate an ASR+LLM cascade using Whisper Large-v3 \cite{radford2023robust} for transcription followed by instruction-tuned large language models (Llama-3 \cite{meta2024llama3}, Mistral-7B \cite{jiang2023mistral7b}, and Qwen2-7B \cite{yang2024qwen2technicalreport}) for downstream reasoning. This establishes a performance ceiling: a significant gap indicates an acoustic processing failure, where low scores in both the ASR and LLM indicate a semantic reasoning failure.

\subsection{Task 2: Child Speech Entailment}

We formalize child speech entailment as a controlled probe for semantic reasoning. Given audio $x$ and a textual hypothesis $h$, the model must determine the logical relationship: $f(x, h) \rightarrow y$, where $y \in \{\textsc{Entailment}, \textsc{Neutral}, \textsc{Contradiction}\}$ \cite{deshmukh}. 

\subsubsection{Hypothesis Generation and Validation}
We generated nine hypotheses per recording (three per entailment class) using LLaMA 3.2 8B. To ensure the integrity of the ground truth, the hypotheses were manually verified by three expert annotators with backgrounds in clinical linguistics and speech-language pathology. These annotators were tasked with validating whether the LLM-assigned labels aligned with the acoustic evidence in the recordings, particularly in cases of dense disfluency.

In total, the annotators manually verified 186 instances (recordings paired with hypotheses). To establish statistical reliability, 50 instances (approx. 27\%) were cross-verified by all three annotators. We measured inter-annotator agreement using Cohen’s Kappa, achieving a score of 0.93. This high level of agreement among annotators validates the reliability of our reasoning probe and ensures that the baseline represents a true semantic ceiling \cite{deshmukh}.

\subsubsection{Difficulty Stratification}
Hypotheses were manually stratified by the annotators during the validation phase to reflect increasing semantic reasoning demands. This stratification was applied within each recording, ensuring that every audio file in the dataset is associated with at least one hypothesis from each of the three difficulty levels. This intra-recording balance allows us to evaluate model performance across varying complexities while keeping the acoustic context constant. The levels are defined as follows:

\begin{itemize}
\item \textbf{Easy}: Directly supported or refuted by a single, clearly identifiable child utterance (e.g., a direct answer to a prompt).
\item \textbf{Medium}: Requires paraphrasing or light inference across nearby utterances, where meaning must be extracted despite local disfluencies.
\item \textbf{Hard}: Requires integrating evidence across multiple child utterances scattered throughout the recording, demanding long-range context retention and resistance to adult-speech interference.
\end{itemize}

\section{Experimental Design and Setup}

Our experimental framework evaluates the transition from modular cascades where audio is first converted to text via ASR before being processed by an LLM to end-to-end audio-semantic reasoning, where a single model extracts meaning directly from raw acoustic signals. This allows us to determine if bypassing the 'transcription bottleneck' enables models to better interpret disfluent speech that traditional ASR systems often mischaracterize.

\subsection{Datasets and Disfluency Profiling}
We evaluate models using the \emph{Voices of Children Who Stutter} corpus \cite{ratner2019}. The dataset comprises 44 total recordings from 22 children, with each participant contributing one sample to the Single-Speaker Reading environment and one to the Mixed-Speaker Interview environment. This setup provides 22 'clean' child-only samples and 22 'noisy' multi-speaker samples for a balanced evaluation of environment-induced performance degradation. Recording durations range from 5 to 10 minutes.
\begin{itemize}
    \item \textbf{Single-Speaker Reading}: Contains only child speech with varying stuttering severity for Child only summarization.
    \item \textbf{Mixed-Speaker Interview}: Semi-structured child-adult interactions without prior diarization or segmentation for child speech entailment.
\end{itemize}

\section{Models and Evaluation Framework}
\label{sec:models}

We evaluate state-of-the-art audio–language models (ALMs) in a zero-shot setting using fixed prompts. Models are compared against cascaded ASR+LLM baselines and transcript-oracle controls to isolate acoustic and semantic reasoning errors.

\subsection{Audio--Language Models}

ALMs couple an audio encoder with a large language model to directly generate responses from raw speech. We evaluate:

\begin{itemize}
    \item \textbf{Audio Flamingo 3 and 2 (7B) \cite{goel2025, ghosh2025}}: pairs a 7B LLM with an AF-Whisper encoder, supporting long-context and chain-of-thought reasoning.
    \item \textbf{Kimi-Audio \cite{kimiteam2025}}: an audio foundation model using a hertz-level tokenizer to discretize continuous audio; the backbone is based on a large-scale multimodal transformer.
    \item \textbf{Qwen2.5-Omni (7B) \cite{yang2024qwen2technicalreport}}: a unified multimodal model with a Thinker–Talker architecture.
    \item \textbf{Qwen2-Audio (7B) \cite{qwen2audio2024}}: simplifies audio pretraining using natural-language prompts.
    \item \textbf{SALMONN (7B) \cite{salmonn2023}}: connects speech/audio encoders with a Vicuna-13B pre-trained LLM.
    \item \textbf{GAMA (7B) \cite{ghosh2024gama}}: integrates multiple audio representations via a Q-Former aggregator into a 7B parameter language model.
    \item \textbf{LTU \cite{gong2024}}: a 7B parameter model trained on OpenAQA-5M for audio question answering with autoregressive reasoning.
\end{itemize}
\subsection{Cascaded ASR + LLM Baselines}

To compare audio-native reasoning with transcript-mediated reasoning, we construct strong cascaded baselines. Audio is first transcribed using:
\textbf{Whisper Large-v3 \cite{radford2023robust}}, 
     \textbf{Granite 3.3.2 \cite{saon_granite-speech_2025}}

Transcripts are then processed by instruction-tuned text LLMs: This setup provides a strong transcript-based upper bound and allows us to disentangle acoustic errors from semantic reasoning failures.

\begin{itemize}
    \item \textbf{Llama 3.2 (8B) \cite{meta2024llama3}}
    \item \textbf{Mistral (7B) \cite{jiang2023mistral7b}}
    \item \textbf{Qwen 2.5 (7B) \cite{bai2023qwen}}
\end{itemize}

\subsection{Evaluation Protocol}

All models are evaluated in a zero-shot setting with identical task prompts. For summarization, outputs are assessed using three independent LLM judges to mitigate evaluator bias and BERTScore F1. For entailment, we report macro-averaged Accuracy and F1 scores, stratified by hypothesis difficulty.

\section{Results}

\subsection{Summarization performance}
\label{sec:results:summ}

Table~\ref{tab:summarization_results} summarises the mean and standard deviation of the five–point ratings (Fluency \cite{fabbri_summeval_2021}, Coherence \cite{maynez_faithfulness_2020}, Faithfulness \cite{maynez_faithfulness_2020}, Relevance \cite{fabbri_summeval_2021}, Purity\cite{anguera_purity_2006})  assigned by three independent LLM judges (Llama~3.2, Mistral~7B and Qwen~2.5) for the interview Summarization task. Two models stand out: \textsc{AudioFlamingo3} and \textsc{Kimi} both achieve mean overall scores above~3, with \textsc{AudioFlamingo3} delivering exceptionally fluent and pure summaries (mean Fluency~4.28 and Purity~4.73) at the cost of slightly reduced faithfulness. The baseline systems \textsc{SALMONN}, \textsc{Qwen 2 audio} and \textsc{GAMA} lag significantly, with overall scores below~2.6. These relative rankings are consistent across the individual sub‑metrics and across judges.

To assess semantic fidelity quantitatively we compute BERTScore~F1 between each audio‑model summary and a transcript‑oracle summary obtained from the human transcript.  The results, shown in Table~\ref{tab:bertscore_results}, largely mirror the subjective ratings: \textsc{Kimi} attains the highest F1 on interview recordings (0.478), narrowly ahead of \textsc{AudioFlamingo3}. Low scores for \textsc{SALMONN} and \textsc{GAMA} corroborate their poor judged performance. Together, these findings indicate that contemporary ALMs can generate coherent child‑centric summaries, but only a subset maintain strong semantic fidelity in the presence of disfluency and adult interference.

\begin{table}[htbp]
\centering
\caption{Mean LLM‑judge scores for \textsc{Interview} summaries averaged across three judges (mean~±~SD; higher is better). Best per column in \textbf{bold}.}
\label{tab:summarization_results}

\resizebox{\columnwidth}{!}{
\begin{tabular}{lcccc} 
\toprule
\textbf{Model} & \textbf{Overall} & \textbf{Fluency} & \textbf{Faithfulness} & \textbf{Purity} \\ 
\midrule

Audio Flamingo 3 & \textbf{3.43 (±0.39)} & \textbf{4.28 (±0.30)} & \textbf{3.07 (±0.30)} & \textbf{4.73 (±0.10)} \\
Audio Flamingo 2 & 2.87 (±0.50) & 3.90 (±0.55) & 2.20 (±0.45) & 4.60 (±0.12) \\
SALMONN 7B & 2.47 (±0.69) & 3.55 (±0.50) & 1.87 (±0.48) & 4.38 (±0.30) \\
Qwen 2 audio 7B & 2.54 (±0.59) & 3.96 (±0.69) & 1.85 (±0.52) & 4.64 (±0.09) \\
GAMA 7B & 2.10 (±0.65) & 2.51 (±1.08) & 1.33 (±0.47) & 4.11 (±0.11) \\

Kimi audio & 3.13 (±0.44) & 4.11 (±0.45) & 2.67 (±0.50) & 4.48 (±0.14) \\
\bottomrule
\end{tabular}}
\end{table}

\begin{table}[htbp]
\centering
\caption{
BERTScore F1 performance across tasks (higher is better, lower is worse). 
Values show mean and 95\% confidence intervals.
}
\footnotesize
\label{tab:bertscore_results}
\resizebox{\columnwidth}{!}{
\begin{tabular}{lcc}
\toprule
\textbf{Model} & \textbf{Interview F1 [95\% CI]} & \textbf{Reading F1 [95\% CI]} \\
\midrule
Audio Flamingo 3 & 0.433 [0.202, 0.262] & \textbf{0.504 [0.180, 0.229]} \\
Qwen 2 audio 7B & 0.294 [0.161, 0.226] & 0.210 [0.143, 0.272] \\
Kimi-audio & \textbf{0.478 [0.248, 0.311]} & 0.209 [0.146, 0.290] \\
SALMONN & 0.188 [0.156, 0.218] & 0.299 [0.217, 0.383] \\
GAMA & 0.053 [0.036, 0.068] & 0.021 [0.000, 0.040] \\
\bottomrule
\end{tabular}}
\end{table}

\subsection{Entailment and difficulty performance}\label{sec:results:analysis}
Table~\ref{tab:entailment_overall} reports the overall accuracy and macro‑averaged F1 scores for the interview entailment task.  \textsc{Qwen2.5‑Omni} achieves the best overall performance (0.681 accuracy, 0.683 F1) and maintains high entailment accuracy (E‑ACC), whereas \textsc{Kimi} attains the highest contradiction accuracy (C‑ACC = 0.811) despite a slightly lower overall score.  Other ALMs trail considerably: \textsc{Qwen2 audio}, \textsc{SALMONN}, \textsc{AudioFlamingo2/3}, \textsc{LTU} and \textsc{GAMA} rarely predict the contradiction class (C‑ACC$<$0.12).  These results reveal a pronounced model‑wise bias toward predicting entailment, which inflates overall accuracy and masks poor performance on challenging examples.  

\begin{table}[t]
\centering
\caption{Audio entailment performance averaged across difficulty levels.
Best values are shown in \textbf{bold}.}
\label{tab:entailment_overall}
\footnotesize
\begin{tabular}{l c c c c}
\toprule
\textbf{Model} & \textbf{ACC} & \textbf{F1} & \textbf{E-ACC} & \textbf{C-ACC} \\
\midrule
Qwen2.5-Omni & \textbf{0.681} & \textbf{0.683} & 0.812 & 0.550 \\
Kimi-Audio & 0.647 & 0.582 & 0.754 & \textbf{0.811} \\
Qwen2-Audio-7B & 0.594 & 0.601 & 0.522 & 0.579 \\
SALMONN & 0.411 & 0.315 & 0.304 & 0.000 \\
Audio Flamingo-3 (v3) & 0.386 & 0.284 & \textbf{0.971} & 0.119 \\
Audio Flamingo-3 (base) & 0.377 & 0.255 & \textbf{1.000} & 0.086 \\
LTU & 0.367 & 0.228 & 0.449 & 0.000 \\
Audio Flamingo-2 & 0.304 & 0.142 & 0.058 & 0.000 \\
GAMA & 0.275 & 0.199 & 0.435 & 0.000 \\
\bottomrule
\end{tabular}
\end{table}

\begin{table}[t]
\centering
\caption{Audio entailment performance averaged across all models by difficulty level.}
\label{tab:entailment_difficulty_avg}
\footnotesize
\begin{tabular}{lccccc}
\toprule
\textbf{Difficulty} & \textbf{ACC} & \textbf{F1} & \textbf{E-ACC} & \textbf{C-ACC} & \textbf{N-ACC} \\
\midrule
Easy   & 0.449 & 0.343 & 0.700 & 0.216 & 0.442 \\
Medium & 0.439 & 0.357 & 0.693 & 0.289 & 0.454 \\
Hard   & 0.417 & 0.327 & 0.644 & 0.273 & 0.501 \\
\bottomrule
\end{tabular}
\end{table}


Across models we also examine performance by difficulty (Table~\ref{tab:entailment_difficulty_avg}).  Overall accuracy decreases only modestly from easy to hard hypotheses (0.449\,$\rightarrow$\,0.417) while contradiction accuracy remains uniformly low (below~0.29).  These patterns indicate that the systematic bias toward the entailment class persists irrespective of reasoning complexity and not because harder items inherently degrade performance.

\subsection{Prompt analysis}
 Prompt engineering yields only modest improvements, with gains largely confined to medium‑difficulty examples, and does not significantly improve performance on easy or hard cases.

Table~\ref{tab:entailment_prompt_comparison} summarizes the final evaluation of the different prompt variants tested with AudioFlamingo3 on the child speech entailment task.  Overall performance differences across prompt designs were modest: the reasoning-based prompt (v3) and the simplified direct prompt (v4) achieved the highest overall accuracies (0.386) compared with 0.377 for the original prompt and 0.367 for the few-shot variant (v2).  The largest single–difficulty improvement occurs on medium‑difficulty hypotheses, where the reasoning prompt attains 0.435 accuracy and 0.356 macro~F1.

We explored four prompt designs for the entailment task using \textsc{AudioFlamingo3}—the original (zero‑shot) formulation, a few‑shot variant, a reasoning‑focused version (v3), and a simple direct instruction (v4).  As shown in Table~\ref{tab:entailment_prompt_comparison}, the reasoning and simple variants achieve small but consistent improvements in overall accuracy and macro F1 (up to about 2--3 percentage points) relative to the baseline.  These gains are concentrated on easy and medium‑difficulty hypotheses; easy and hard items see little change.  Crucially, all prompts still produce overwhelmingly entailment predictions (87--92\% of outputs) and almost never predict contradiction, indicating that prompt engineering cannot by itself remedy the class imbalance.

\begin{table}[t]
\centering
\caption{Audio entailment performance by prompt version and difficulty.
Metrics are macro-averaged across entailment, neutral, and contradiction classes.}
\label{tab:entailment_prompt_comparison}
\footnotesize
\begin{tabular}{l l c c c c}
\toprule
\textbf{Version} & \textbf{Difficulty} & \textbf{ACC} & \textbf{F1} & \textbf{P} & \textbf{R} \\
\midrule
Original & Easy   & 0.406 & 0.305 & 0.542 & 0.405 \\
Original & Medium & 0.362 & 0.230 & 0.451 & 0.364 \\
Original & Hard   & 0.362 & 0.229 & 0.536 & 0.362 \\
\midrule
v2 Few-shot & Easy   & 0.362 & 0.228 & 0.451 & 0.362 \\
v2 Few-shot & Medium & 0.391 & 0.300 & 0.672 & 0.392 \\
v2 Few-shot & Hard   & 0.348 & 0.181 & 0.409 & 0.261 \\
\midrule
v3 Reasoning & Easy   & 0.377 & 0.254 & 0.620 & 0.378 \\
v3 Reasoning & Medium & \textbf{0.435} & \textbf{0.356} & 0.597 & \textbf{0.437} \\
v3 Reasoning & Hard   & 0.348 & 0.242 & 0.513 & 0.348 \\
\midrule
v4 Simple & Easy   & 0.391 & 0.214 & 0.431 & 0.294 \\
v4 Simple & Medium & 0.406 & 0.323 & 0.565 & 0.408 \\
v4 Simple & Hard   & 0.362 & 0.186 & 0.439 & 0.272 \\
\bottomrule
\end{tabular}
\end{table}

\begin{table}[t]
\centering
\caption{Child speech entailment cascade results with abbreviated ASR (G: Granite, W: Whisper) and LLM (L: Llama, M: Mistral, Q: Qwen). Overall metrics (N = 207).}
\label{tab:interview_cascade_overall}
\footnotesize
\begin{tabular}{llccccc}
\toprule
\textbf{ASR} & \textbf{LLM} & \textbf{ACC} & \textbf{F1} & \textbf{E-ACC} & \textbf{N-ACC} & \textbf{C-ACC} \\
\midrule
G  & L & 0.478 & 0.427 & 0.942 & 0.087 & 0.406 \\
G  & M & 0.425 & 0.345 & 0.145 & 0.986 & 0.145 \\
G  & Q & 0.715 & 0.715 & 0.710 & 0.623 & 0.812 \\
W  & L & 0.560 & 0.537 & 0.913 & 0.188 & 0.580 \\
W  & M & 0.469 & 0.422 & 0.333 & 0.928 & 0.145 \\
W  & Q & \textbf{0.739} & \textbf{0.737} & 0.870 & 0.536 & \textbf{0.812} \\
\bottomrule
\end{tabular}
\end{table}

\begin{table}[t]
\centering
\caption{Child speech entailment results by hypothesis difficulty (N=69 per level). Abbreviations: ASR (G~$=$~Granite, W~$=$~Whisper) and LLM (L~$=$~Llama, M~$=$~Mistral, Q~$=$~Qwen).}
\label{tab:interview_cascade_by_difficulty}
\footnotesize
\resizebox{\columnwidth}{!}{%
\begin{tabular}{llcccccc} 
\toprule
& & \multicolumn{2}{c}{\textbf{Easy}} 
& \multicolumn{2}{c}{\textbf{Medium}} 
& \multicolumn{2}{c}{\textbf{Hard}} \\
\cmidrule(lr){3-4} \cmidrule(lr){5-6} \cmidrule(lr){7-8} 
\textbf{ASR} & \textbf{LLM} 
& ACC & F1  
& ACC & F1  
& ACC & F1 \\ 
\midrule
G & L & 0.478 & 0.436 & 0.536 & 0.492 & 0.420 & 0.342 \\
G & M & 0.449 & 0.391 & 0.464 & 0.382 & 0.362 & 0.246 \\
G & Q & 0.725 & 0.726 & 0.667 & 0.667 & 0.754 & 0.753 \\
W & L & 0.565 & 0.542 & 0.594 & 0.571 & 0.522 & 0.495 \\
W & M & 0.449 & 0.412 & 0.435 & 0.361 & 0.522 & 0.485 \\
W & Q & \textbf{0.812} & \textbf{0.806} & 0.681 & 0.679 & 0.725 & 0.726 \\
\bottomrule
\end{tabular}}
\end{table}

\begin{table}[t]
\centering
\footnotesize
\caption{Mean performance by disfluency density quantile.  Columns show the density range, number of interview audios ($n$), mean density, mean entailment accuracy (ACC) on the child speech entailment and mean Summarization faithfulness (Faith).}
\footnotesize
\label{tab:density_analysis}
\begin{tabular}{lccccc}
\toprule
\textbf{Density bin} & \textbf{$n$} & \textbf{Density} & \textbf{ACC} & \textbf{Faith} \\
\midrule
(0.0115, 0.0356] & 7 & 0.023 & 0.542 & 2.49 \\
(0.0356, 0.0709] & 6 & 0.047 & 0.567 & 1.86 \\
(0.0709, 0.105] & 6 & 0.086 & 0.662 & 2.39 \\
(0.105, 0.332] & 7 & 0.170 & 0.529 & 2.34 \\
\bottomrule
\end{tabular}
\end{table}

\begin{table}[!h]
\centering
\caption{Mean performance by dominant disfluency type.  Columns list the disfluency type, number of audios ($n$), mean disfluency density, mean entailment accuracy (ACC) and mean Summarization faithfulness (Faith).}
\label{tab:type_analysis}
\footnotesize
\begin{tabular}{lccccc}
\toprule
\textbf{Disfluency type} & \textbf{$n$} & \textbf{Density} & \textbf{ACC} & \textbf{Faith} \\
\midrule
filled\_pause & 18 & 0.078 & 0.574 & 2.22 \\
mixed & 6 & 0.117 & 0.559 & 2.40 \\
repetition & 2 & 0.021 & 0.500 & 2.50 \\
\bottomrule
\end{tabular}
\end{table}

Across all ASR+LLM cascades (Tables~\ref{tab:interview_cascade_overall} and~\ref{tab:interview_cascade_by_difficulty}) we observe that pairing Whisper with Qwen yields the strongest performance: it achieves 0.739 overall accuracy and 0.737 macro F1 and maintains high contradiction accuracy across difficulty levels.  Granite combined with Qwen is the second‑best cascade. In contrast, cascades using Llama or Mistral for the reasoning stage underperform because they rarely predict contradictions even when transcripts are accurate.  These results suggest that current end‑to‑end audio models struggle primarily with acoustic processing in the presence of stuttering and interleaved speakers when supplied with accurate transcripts the underlying text LLMs can achieve balanced entailment and contradiction detection.

\subsection{Error analysis: Disfluency density and type effects}

We examine whether increased disfluency burden in child speech correlates with degraded reasoning and Summarization performance. For each interview recording, we extract counts of repetitions ($N=2$), prolongations, blocks (defined as silent postural fixations or audible airflow stoppages), and filled pauses ($N=18$). These counts were derived from the ground-truth manual annotations provided in the Voices-CWS corpus, ensuring that our density metrics reflect clinical reality rather than ASR transcription errors. We compute the total disfluency rate as the ratio of these markers to the total word count.

Recordings are binned into four quantile ranges by disfluency density; within each bin, we measure mean entailment accuracy and mean Summarization faithfulness (Table~\ref{tab:density_analysis}). As shown, performance remains relatively stable for low to moderate densities but drops in the highest-density bin. Faithfulness varies more widely across bins, suggesting that Summarization quality is more sensitive to the structural disruption caused by disfluency than the three-class entailment task.

We also group recordings by the dominant disfluency type: filled pauses, mixed ($N=6$), or repetitions. We then summarise their mean density, entailment accuracy, and faithfulness (Table~\ref{tab:type_analysis}). Most interviews in this dataset are filled-pause dominated. While the mixed and repetition cases show slightly lower entailment accuracy, these trends should be interpreted cautiously due to the small sample size in the non-pause categories.

\section{Conclusion}
This paper evaluated state-of-the-art audio language models (ALMs) on summarization and semantic reasoning in disfluent child speech. Across models, we observe a pronounced bias toward entailment predictions, with contradiction detection remaining weak even when overall accuracy appears high. Difficulty-stratified results show that this bias persists across easy, medium, and hard hypotheses, suggesting that performance limitations stem less from reasoning complexity and more from semantic calibration under noisy acoustic input.

\section{Use of Generative AI Disclosure}
The authors used generative AI tools (specifically ChatGPT) solely for grammatical refinement, structural editing, and LaTeX formatting of the manuscript. The authors reviewed and edited all AI-generated suggestions and take full responsibility for the technical accuracy and integrity of the final content.

\bibliographystyle{IEEEtran}
\bibliography{clean_references}

\end{document}